\documentclass[sigconf, noacm]{acmart}

\usepackage{listings}
\usepackage[scale=0.85]{sourcecodepro}
\usepackage[T1]{fontenc}

\lstdefinelanguage{none}{
  identifierstyle=
}

\AtBeginDocument{%
  \providecommand\BibTeX{{%
    \normalfont B\kern-0.5em{\scshape i\kern-0.25em b}\kern-0.8em\TeX}}}

\setcopyright{rightsretained}
\copyrightyear{2022}
\acmYear{2022}
\acmConference[NEC SX-Aurora Articles]{NEC SX-Aurora Articles}{May 2022}{}

\begin{document}

\title{SOL: Reducing the Maintenance Overhead for Integrating Hardware Support into AI Frameworks}

\author{Nicolas Weber}
\email{nicolas.weber@neclab.eu}
\affiliation{%
	\institution{NEC Laboratories Europe}
}

\begin{abstract}
The increased interest in Artificial Intelligence (AI) raised the need for
highly optimized and sophisticated AI frameworks. Starting with the Lua-based
Torch many frameworks have emerged over time, such as Theano~\cite{THEANO},
Caffe~\cite{CAFFE}, Chainer~\cite{CHAINER}, CNTK~\cite{CNTK},
MxNet~\cite{MXNET}, PyTorch~\cite{PYTORCH}, DL4J, or
TensorFlow~\cite{TENSORFLOW}.

All of these provide a high level scripting API that allows users to easily
design neural networks and run these on various kinds of hardware. What the user
usually does not see is the high effort put into these frameworks to provide
peak execution performance.

While mainstream CPUs and GPUs have the "luxury" to have a wide spread user base
in the open source community, less mainstream CPU, GPU or accelerator vendors
need to put in a high effort to get their hardware supported by these
frameworks. This includes not only the development of highly efficient compute
libraries such as CUDNN, OneDNN or VEDNN but also supporting an ever growing
number of simpler compute operations such as summation and multiplications. Each
of these frameworks, nowadays, supports several hundred of unique operations,
with tensors of various sizes, shapes and data types, which end up in thousands
of compute kernels required for each device type. And the number of operations
keeps increasing.

That is why NEC Laboratories
Europe\footnote{\href{https://www.neclab.eu}{www.neclab.eu}} started developing the
SOL\footnote{\href{https://sol.neclab.eu}{sol.neclab.eu}; "SOL" is no
acronym and means "sun" in many languages.} AI Optimization project already
years ago, to deliver optimal performance to users while keeping the maintenance
burden minimal.
\end{abstract}


\maketitle

\section{AI Framework Execution Phases}
The key insight of SOL is based on understanding the execution phases of AI
frameworks which are outlined in the following code example Listing~\ref{lst1}.

\begin{lstlisting}[caption={Simplistic PyTorch code that initializes a model, generates input data and runs it in inference mode.}, float=t, label=lst1]
import torch
import torchvision

# 1. Model Initialization Phase
model = torchvision.resnet50()
model = model.to(device)

# 2. Data Preparation Phase
input = torch.rand(1, 2, 3)

# 3. Model Execution Phase
with torch.no_grad():
	output = model(input)
\end{lstlisting}

The first phase is the "Model Initialization". In this phase the neural network
structure gets defined and model weights either get loaded from a pre-trained
dataset or randomly initialized. Depending on the framework, the model either
gets directly initialized on the target device or it needs to be manually
transferred.

In the second phase, the inference or training datasets get processed. This
often is done by the Numpy, SciKit Learn or Pandas libraries.

Last, the model gets executed either in a simple inference task or in a training
loop. Let's take a closer look at a training loop (Listing~\ref{lst2})

\begin{lstlisting}[caption={Simple example of a PyTorch training loop.}, float=t, label=lst2]
optimizer		= torch.optim.sgd(
	model.parameters())
loss_function	= torch.nn.functional.l1_loss
for epoch in range(epochs):
	for input, target in dataset:
		output = model(input)
		loss   = loss_function(output, target)
		loss.backward()
	optimizer.step()
\end{lstlisting}

First, we build up an optimizer that takes care of updating our model weights
during training and a loss function. We then have multiple nested loops that
iterate over the batches of our dataset and multiple epochs. Within our loop we
execute the forward execution of our model, compute the loss and execute the
backward pass, which computes the gradients. Last, we call the optimizer to
update our weights.

\section{SOL principle}
\begin{figure*}[t]
	\includegraphics[width=\linewidth, clip, trim=70 120 120 235]{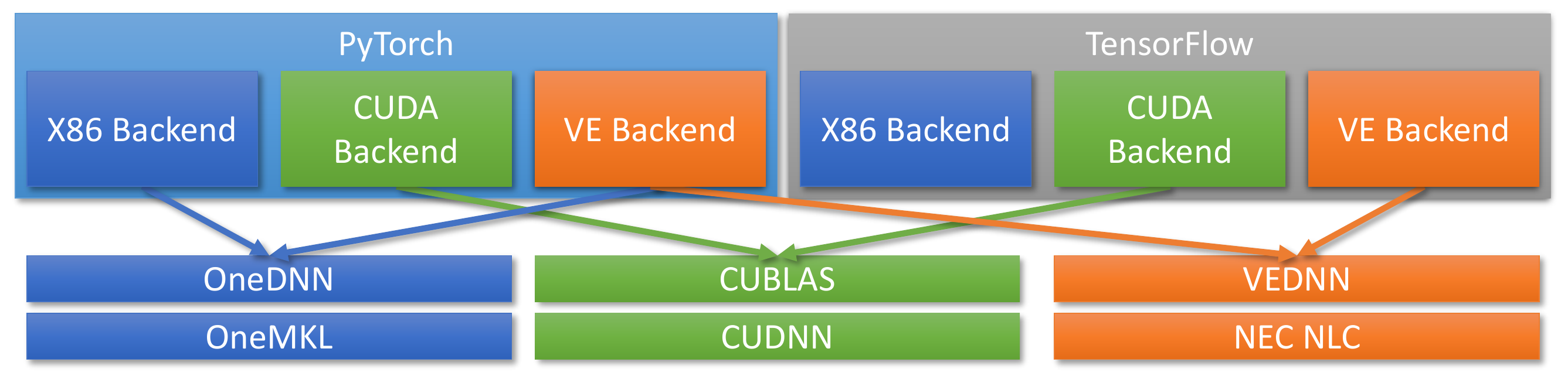}
	\caption{State-of-the-art device support in AI frameworks. Hardware vendors
	upstream the entire device support into the codebase of the AI framework and
	only share external compute libraries for compute heavy kernels, i.e.,
	Convolution or GEMM.}
	\label{fig1}
\end{figure*}

What hardware vendors so far have been doing was to port all low level function
calls of AI frameworks to their own compute libraries, which then directly can
be executed by the AI framework. This is a huge effort and requires lots of
workforce to implement and maintain all of these functions (Figure~\ref{fig1}).

When looking at the training loop (Listing~\ref{lst2}), we see that we have four atomic steps.

\begin{enumerate}
\item the forward pass of the model
\item the execution of the loss function
\item the backward pass of the model
\item the update of the model weights
\end{enumerate}

Independently from the used AI framework these four steps get triggered within
training loops. Which means that it does not matter if the framework calls a
single compute kernel within these steps, or hundreds of these. The framework
expects the output after the step, but how this output is computed is not
defined.

\begin{figure}[t]
	\includegraphics[width=\columnwidth, clip, trim=70 250 480 40]{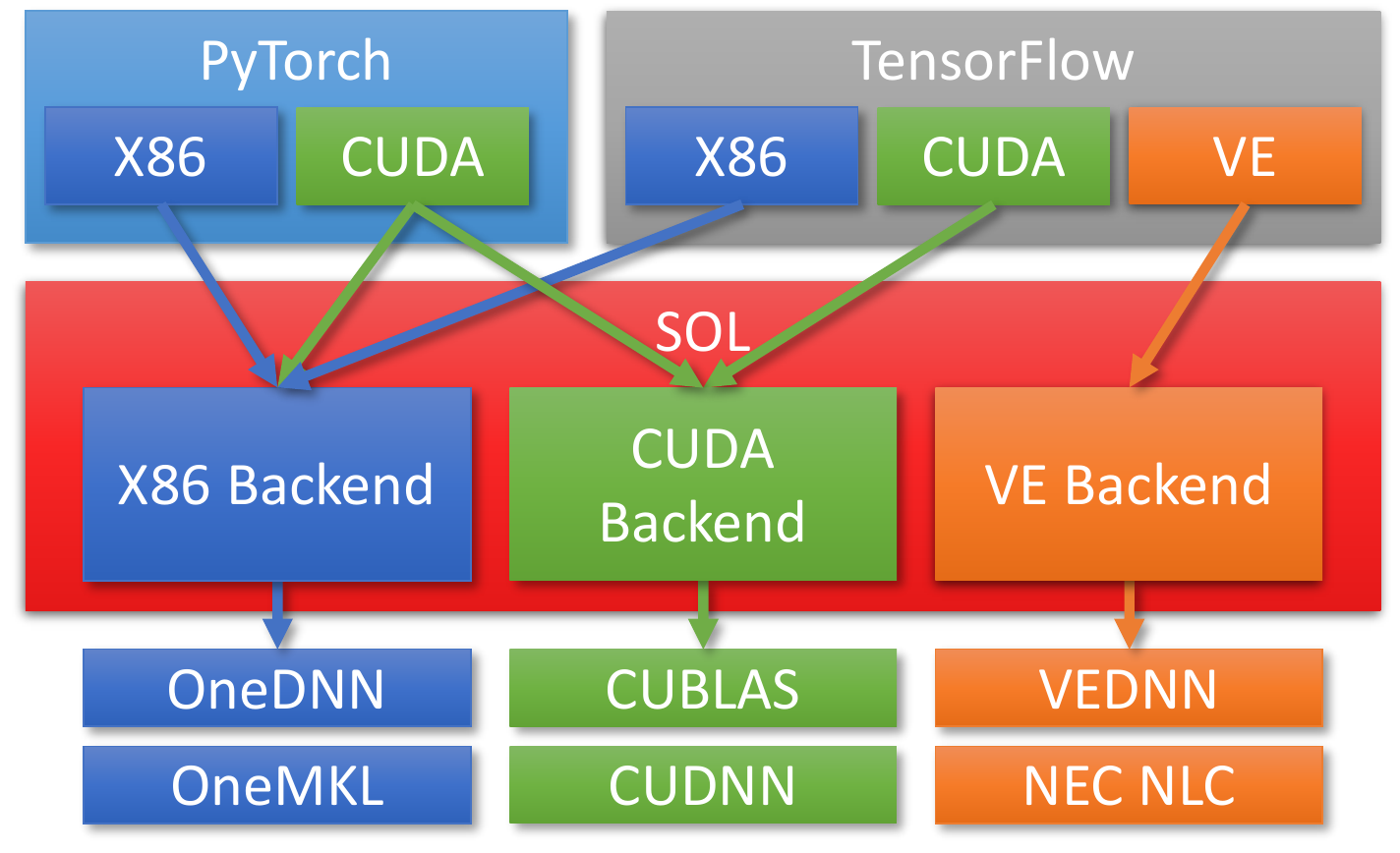}
	\caption{In contrast to the state-of-the-art, SOL requires only minimal
	device support within the AI frameworks and shares the entire optimization
	and execution functionalities across all AI frameworks. For frameworks
	without support to add native device support, SOL can use it's transparent
	offloading technique to still run computations on accelerators.}
	\label{fig2}
\end{figure}

Therefore, within SOL we replace the 1st and 3rd step with our neural network
compiler engine. For the user this just means to run a single command which
returns a new instance of the neural network, that is fully compatible with the
AI framework, but whose computations do rely on SOL instead of the AI
framework's execution engine. With this trick, a hardware vendor does not need
to support over 90\% of the AI framework's compute kernels, as for this we use
our unified compiler engine that can be utilized in any AI frameworks with
little effort. So, instead of constantly maintaining compute kernels for each AI
framework separately, and keep up with the fast development cycles and API
changes, SOL only requires a single unified device support and abstracts the
entire integration into the AI framework (Figure~\ref{fig2}).

This leaves us with the loss functions and the weight update, which are mainly
based on very simple computations and are little effort to be integrated into
each of these frameworks.

\begin{lstlisting}[caption={Minimal steps to apply SOL optimizations.}]
import sol
sol_model = sol.optimize(model, ...)
\end{lstlisting}

To add this minimal necessary device support of the NEC SX-Aurora TSUBASA to
TensorFlow and PyTorch, we developed the veda-tensorflow~\cite{VEDATF} and
veda-pytorch~\cite{VEDAPYTORCH} Python modules. These only contain these minimal
required compute kernels and necessary callbacks for the AI frameworks.
veda-tensorflow is based on the TensorFlow PluggableDevice API~\cite{TFPLUG}
that has been introduced in TensorFlow v2.6. For PyTorch we use their function
registry to side-load our compute kernels into PyTorch's execution engine and
added the "VE" device type in PyTorch v1.10. This allows to add the entire
SX-Aurora device support without any code changes to TensorFlow or PyTorch.

Frameworks that don't have the option to register device support at runtime
(i.e. DL4J) still can be used with SOL through our transparent
offloading~\cite{HPML} mechanism. With transparent offloading, SOL is able to
use a model that is stored in the host memory and execute it on any target
device. For this, SOL maintains a copy of the model on the device and
transparently synchronizes changes to the weights between the host and the
device memory. However, this method can have performance penalties due to memory
synchronization and should not be used when
native hardware integration is available.

\section{SOL architecture}

SOL uses an extendable plugin infrastructure. Heart of SOL is the so called
"core" that contains most of it's functionality, graph representation and
transformations. Further, SOL consists of:

\begin{description}
\item[JITs:] plugins that enable to compile generated source code using external
  compilers such as GCC, LLVM or ISPC.
\item[Frameworks:] plugins that serve three purposes:
	\begin{enumerate}
		\item parsing the AI framework specific neural network format and translate them to SOL's own *High Level Intermediate Representation* (HLIR)
		\item generating wrappers that allow to integrate SOL optimized neural network's into the AI frameworks
		\item connect SOL's runtime to the AI framework's memory allocation system
	\end{enumerate}
\item[Devices:] plugins providing basic device functionality like enumeration, memory management, etc.
\item[Backends:] plugins that translate parts of the HLIR graph to device specific and optimized implementations.
\end{description}

When \texttt{sol.optimize(...)} is called, SOL performs the following steps.
\begin{enumerate}
\item the framework plugin parses the neural network structure and translates it
   into SOL's HLIR format.
\item SOL applies generic optimizations and transformations, i.e., removing layers
   that do not contribute to any output of the neural networks, are
   mathematically irrelevant, or can be statically evaluated because their input
   is constant.
\item generates/compiles framework specific wrappers that integrate the optimized
   model into the AI framework
\item return an instance of this optimized model to the user.
\end{enumerate}

When the model gets executed the very first time, SOL generates three versions
of the neural network.
\begin{enumerate}
\item an inference version
\item a training forward version
\item a training backward version
\end{enumerate}

Each of these versions get optimized individually. During the optimization
procedure, SOL performs a layer-by-layer auto-tuning and selects the backends
that perform best. SOL allows that multiple backends (i.e., different BLAS
libraries) implement the same layers, and the best performing gets chosen. In
contrast to tools like TVM~\cite{TVM} that perform an exhaustive and very long
auto-tuning (up to several hours), SOL relies on a layer-by-layer auto-tuning,
that is usually completed within seconds.

After determining the best backend for each layer, SOL groups connected layers
that have the same backend assigned to enable layer merging. All of the grouped
layers get compiled by the respective backends and finally get linked to a
single library that contains the entire execution pipeline and compute kernels
for the respective neural network.

Since SOL v0.4.1 we support to extend SOL through its SDK. SOL's
plugin-infrastructure makes it easy extendable. All parts of SOL can be
extended, the backends, jits, frameworks and devices. Further, entirely new
layers and transformation passes can be added. Please refer to the
SDK\footnote{\href{https://sol.neclab.eu/docs/v0.5.0/sdk.html}{sol.neclab.eu/docs/v0.5.0/sdk.html}} for more information.

\section{SOL Benefits}
So far we mainly discussed how the concept behind SOL can reduce the maintenance
effort for hardware vendors to support new hardware within AI frameworks with
little effort. But SOL also provides many benefits to AI users through its
unique design principles.

\subsection{Performance}
Performance is a key aspect in AI training as due to the high computational
demands the costs of training models can be extreme, i.e., 12M\$ for training
GPT-3~\cite{VBEAT}.

Therefore, a lot of work is invested in making the AI frameworks as efficient as
possible. But still, depending on which neural network you are running, the one
or other AI framework might be faster.

\begin{figure}[t]
	\includegraphics[width=\columnwidth, clip, trim=155 10 155 10]{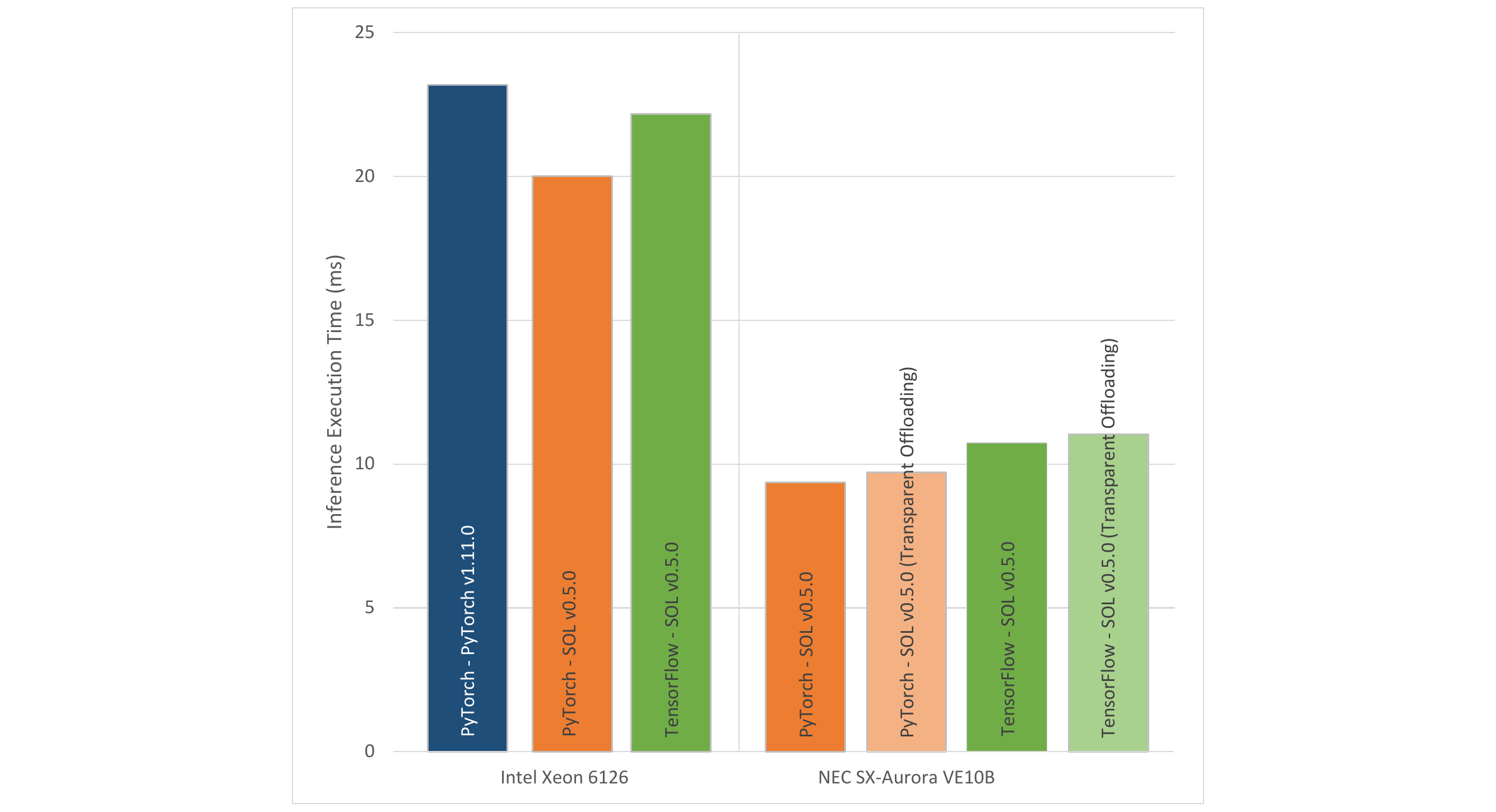}
	\caption{Inference performance comparison of SOL of the TorchVision
		Resnet50, executed in vanilla PyTorch v1.11.0, PyTorch + SOL and
		cross-executing the PyTorch defined model in TensorFlow v2.8.0 + SOL
		(see next section for details). You can see that on Intel Xeon 6126 SOL
		is ~15\% faster than vanilla PyTorch. Further, TensorFlow has a higher
		execution overhead compared to PyTorch when running the identical SOL
		model. When using the NEC SX-Aurora TSUBASA VE10B, we achieve more than
		50\% speedup compared to vanilla PyTorch on the Intel Xeon. As
		previously mentioned transparent offloading inherits a small performance
		penalty compared to native execution because of the memory transfer to
		the accelerator.}
	\label{fig3}
\end{figure}

With SOL this problem does not exist, as it compiles hardware specialized
implementations of the neural networks, which not only outperform the AI
framework's execution engines but also provides equivalent performance
independent of which AI framework is used (Figure~\ref{fig3}). This relieves users from evaluating
which AI framework performs fastest for them and allows them to use the
framework they personally prefer.

\subsection{Cross-Framework Execution}
A common problem for AI developers is that neural networks are written in a
specific framework and it's tedious to convert it to other frameworks.
ONNX~\cite{ONNX} has established itself as *THE* neural network storage format.
However, as of today all major AI frameworks can only export ONNX models and not
load them. Projects like ONNX-TF~\cite{ONNXTF} or
ONNX-Pytorch~\cite{ONNXPYTORCH} try to add this missing functionality but they
often hit limitations of the frameworks themselves, as can be seen at the list
of supported
layers\footnote{\href{https://github.com/onnx/onnx-tensorflow/blob/main/doc/support_status.md}{github.com/onnx/onnx-tensorflow/blob/main/doc/support\_status.md}}.
This is caused by the fact that not all layers, hyper-parameters and data types
supported by PyTorch are available in TensorFlow and vice versa. For example
PyTorch's \textit{AdaptiveAvgPooling} allows arbitrary output shapes, while
TensorFlow's \textit{GlobalAvgPooling} only supports to entirely reduce
the pixel dimensions. Or TensorFlow's CumSum allows inclusive/exclusive and
reversed modes, while PyTorch only supports inclusive CumSums.

This requires teams to agree on a common AI framework to be used and even
manually port (or even adjust) neural networks if they are only available in
foreign formats.

As SOL does not use the AI framework's execution engines, it does not have these
limitations. Therefore SOL can directly execute a PyTorch model within
TensorFlow or vice versa, even for operations that are not natively supported
within TensorFlow.

For users, this only requires to add a single keyword.

\begin{lstlisting}[caption={Minimal changes needed to run any PyTorch model in TensorFlow.}]
sol_model = sol.optimize(pytorch_model, ...,
	framework='tensorflow')
\end{lstlisting}

So far SOL supports loading any Python based neural networks (PyTorch,
TensorFlow or ONNX) and run these in PyTorch, TensorFlow (as \texttt{tf.Module}
or \texttt{tf.keras.model.Model}) or as a plain Python function that uses Numpy
arrays as input. When executing with Numpy arrays the user can use transparent
offloading to execute the neural network on any device and is not bound to the
host CPU.

\subsection{Dynamic Dimensions}
Dynamic dimensions in executing neural networks is a two sided blade. From a
user perspective it allows more flexibility. From the framework perspective it
can decrease the performance as it requires to handle more runtime information
and prevents the application of certain compile-time optimizations. However, as
the users demand it today, all AI frameworks support them.

In all AI frameworks, the user needs to manually identify which dimensions
should be dynamic. Further, using dynamic dimensions in AI frameworks can easily
hit implementation limitations. For example the
\texttt{torch.jit.trace(...)}\footnote{\href{https://pytorch.org/docs/stable/generated/torch.jit.trace.html}{pytorch.org/docs/stable/generated/torch.jit.trace.html}}
function for jit compiling a neural network does not allow to use any dynamic
dimensions at all.

Instead SOL uses a new unique dynamic dimensions system. SOL automatically
determines which dimensions can be dynamic based on the structure of the neural
network. For example, if a layers uses a bias with 128 channels, this dimension
is fixed by the structure of the neural network and cannot be dynamically set.
This method does not require any user input. After analyzing the neural network,
SOL reports the possible input and output shapes as in the following example
(Listing~\ref{lst3}).

\begin{lstlisting}[caption={Input and output shapes inferred by SOL after parsing a network. SOL identified four (\#0-\#3) dynamic dimensions.},language=none,label=lst3, float=t]
Inputs:  in_0 [#0, 5, #1, #2, #3]
Outputs: out_0  {
    "A": [#0, 5, #1, #2, #3],
    "B": [#0, 5, #1, 3, 3],
    "C": [#0, 5, #1, 5, 7],
}
\end{lstlisting}

In this example, SOL has identified four dynamic dimensions (\#0-\#3). However, by
default SOL uses the fixed dimensions and requires the user to explicitly enable
dynamic dimensions manually. This guarantees that SOL can optimize all
dimensions that don't need to be dynamic. To enable a dynamic dimension the
\texttt{sol.optimize(..., vdims=[...])} keyword can be used. SOL allows to
enable (\texttt{True}), disable (\texttt{False}) or overwrite (\texttt{integer >
0}) the values used.

\subsection{Memory Consumption Estimation}
Most users of AI frameworks have already encountered the situation where their
training died with the message: \texttt{out of memory}, which can be troublesome
if it occurs after several days of training. The main reason is that while the
number of model parameters is easy to identify, i.e. by TensorFlows
\texttt{model.summary()} method (Listing 6) it is not possible to get a good
estimate of peak memory consumption from any AI framework without running the
model at least once.

\begin{figure*}[t]
\begin{lstlisting}[caption={Output of \texttt{model.summary()} of Alexnet in Keras.}, language=none]
Model: "model"
_________________________________________________________________
Layer (type)                 Output Shape              Param #
=================================================================
input_1 (InputLayer)         [(None, 224, 224, 3)]     0
conv2d (Conv2D)              (None, 56, 56, 64)        23296
max_pooling2d (MaxPooling2D) (None, 27, 27, 64)        0
conv2d_1 (Conv2D)            (None, 27, 27, 192)       307392
max_pooling2d_1 (MaxPooling2 (None, 13, 13, 192)       0
conv2d_2 (Conv2D)            (None, 13, 13, 384)       663936
conv2d_3 (Conv2D)            (None, 13, 13, 256)       884992
conv2d_4 (Conv2D)            (None, 13, 13, 256)       590080
max_pooling2d_2 (MaxPooling2 (None, 6, 6, 256)         0
flatten (Flatten)            (None, 9216)              0
dense (Dense)                (None, 4096)              37752832
dense_1 (Dense)              (None, 4096)              16781312
dense_2 (Dense)              (None, 1000)              4097000
=================================================================
Total params: 61,100,840
Trainable params: 61,100,840
Non-trainable params: 0
_________________________________________________________________
\end{lstlisting}
\end{figure*}

For example PyTorch's \texttt{torch.cuda.max\_memory\_allocated()} can be used to
get a rough estimate of the memory consumption. This is because of two main
issues. First, dynamic graphs (as used by PyTorch) get evaluated at runtime,
which does not allow to take a grasp of the neural network structure before
executing. And second, during training of neural networks intermediate results
need to be stored between forward and backward pass, which increases the peak
memory consumption during training.

As SOL compiles the neural network and employs a static schedule during this
procedure, it knows exactly when it allocates memory and when it frees it. This
allows SOL to give an estimation of the peak memory consumption before running
for the first time.

\begin{lstlisting}[caption={Peak memory estimated by SOL.}, language=none]
Estimated Peak Memory Consumption:
Inference: ~15MB
Training:  ~25MB
\end{lstlisting}

SOL only reports an estimate, as due to memory alignment and fragmentation it's
impossible to give a more accurate number. SOL can also generate more detailed
reports which show the development of the memory consumption over time and which
kind of data (parameters, inputs, outputs or intermediate) are causing the
memory consumption.

\begin{figure*}[t]
	\fbox{\includegraphics[width=\linewidth, clip, trim=0 0 0 0]{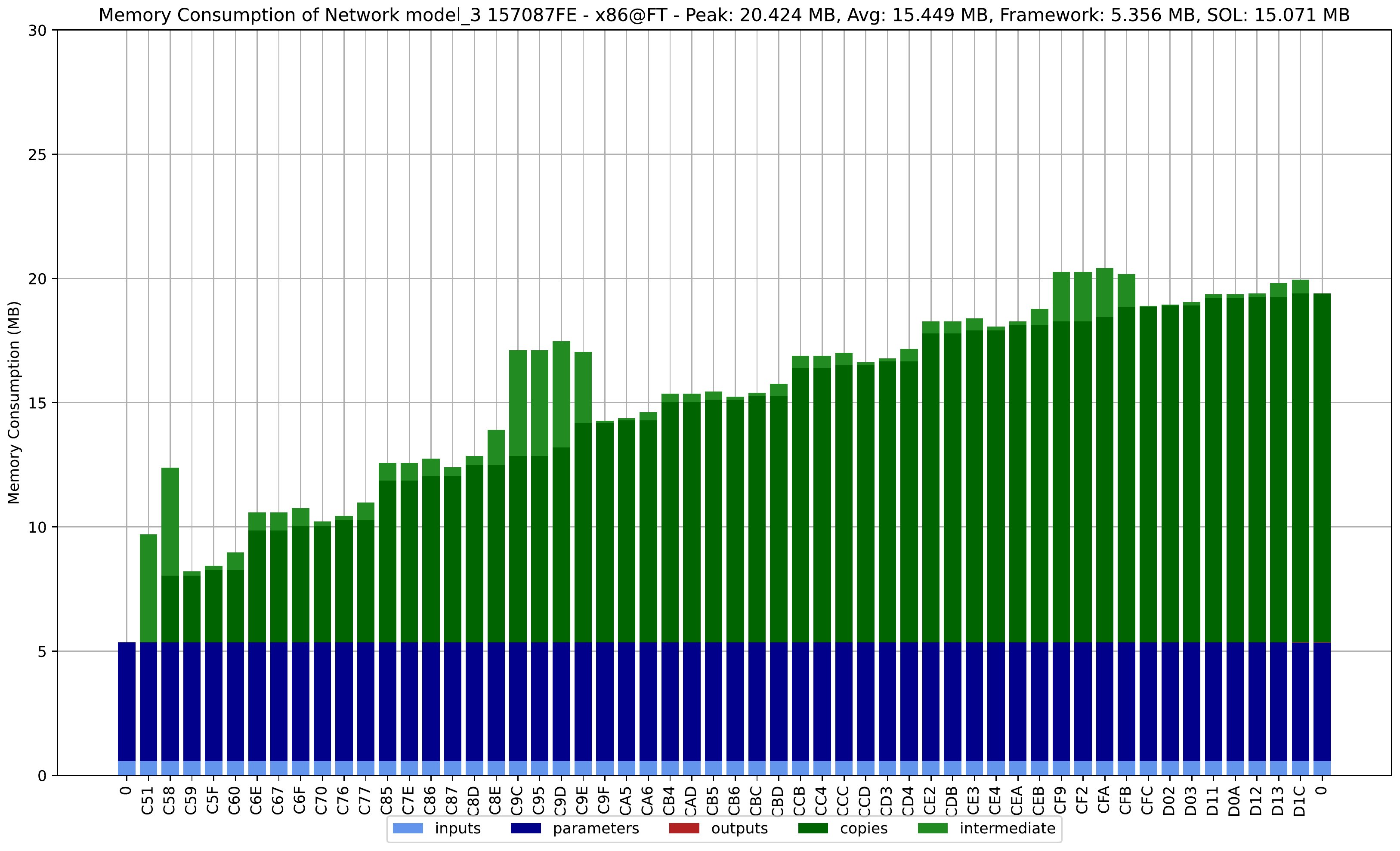}}
	\caption{Advanced SOL memory consumption report for forward pass of
		 training. Each bar does not correspond to a single layer, but to a
		 group of merged layers.}
\end{figure*}

\section{Neural Network Deployment}
After a neural network has been trained, it needs to be deployed in its final
application where it is undesirable to ship the entire AI framework with the
application. There are two types of deployment options available: Runtime
Engines and Compilers.

\begin{itemize}
	\item \textbf{Neural Network Runtime Engines}:
		\begin{itemize}
			\item PyTorch provides \texttt{libtorch} which is a C++ API enabling
to run PyTorch models. However, libtorch with CUDA support is 1.5GB which is
undesirable to be shipped with most applications.
			\item TensorFlow provides TensorFlow-lite which is much more
lightweight but still consumes several hundreds of megabytes.
			\item ONNXRuntime~\cite{ONNXRUNTIME} is a lightweight
	execution engine for ONNX models that supports vendor specific backends.
		\end{itemize}
	\item \textbf{Neural Network Compilers}:
		\begin{itemize}
			\item NVIDIA's TensorRT~\cite{TENSORRT} allows to deploy neural
networks for their own hardware.
			\item Intel's OpenVINO~\cite{OPENVINO} allows deploying
neural networks on all kinds of Intel hardware (CPUs, FPGA, IPU, ...).
			\item TVM is a widely used inference
compiler that supports a huge variety of target platforms.
		\end{itemize}
\end{itemize}

This is only a small list of tools available for deploying neural networks. As
can be seen, especially the compilers (except TVM) are vendor specific and limit
the user to their respective hardware. So if the user needs to support multiple
vendors, it's easier to stick to runtime engines to ease the effort required to
maintain multiple implementations, with the drawback of higher storage
consumption and lower performance compared to compiled neural networks.

Another aspect is the compatibility, especially with the compiler support. When
training your neural network in an AI framework it is not guaranteed that the
compiler you choose is able to compile the given neural networks with the chosen
layers and hyper-parameters.

SOL overcomes these issues as it is not only multi-vendor, but it also
guarantees that if you have trained your neural network with SOL, the exact same
network can also be compiled to an optimized library. As of now, SOL supports
deployment of neural networks in device specific libraries, with minimal storage
footprint, for static and shared linux libraries. In case you need special
adjustments for your system or application, you can also directly use the
optimized source code for your neural network and adjust it to your specific
needs.

\section{Summary}
In this article we presented the SOL AI
Optimization project\footnote{\href{https://sol.neclab.eu}{sol.neclab.eu}}. SOL enhances the usability and performance of AI
frameworks, independently from the used hardware and to enable vendors to add
hardware support easily to any AI framework, with minimal maintenance effort.

If you are interested in using SOL, please apply for
SOL's closed beta\footnote{\href{http://sysml.neclab.eu/projects/sol/closed_beta/}{sysml.neclab.eu/projects/sol/closed\_beta/}}.

\bibliographystyle{ACM-Reference-Format}
\bibliography{main}

\end{document}